\documentclass{article}
\usepackage{spconf,amsmath,graphicx,hyperref}
\usepackage{amsmath, amssymb}
\usepackage{amsmath,amssymb}
\usepackage{algorithm,algorithmic}
\usepackage{amsmath,amssymb}
\usepackage{booktabs}
\usepackage{subcaption}
\usepackage{graphicx}
\usepackage[section]{placeins}

\title{Conformal Signal Temporal Logic for Robust Reinforcement Learning Control: A Case Study
\thanks{Preprint. Accepted at ICASSP 2026.}}
\name{Authors}
\address{Affiliations}
\twoauthors
  {Hani Beirami}
    {
     University of Tehran\\
     hanibeirami@ut.ac.ir}
  {M M Manjurul Islam}
    {
     Ulster University\\
     m.islam@ulster.ac.uk}

\begin{document}
%
\maketitle
\begingroup
\renewcommand\thefootnote{}
\footnotetext{
© 2026 IEEE. Personal use of this material is permitted. Permission from IEEE must be obtained for all other uses, in any current or future media, including reprinting/republishing this material for advertising or promotional purposes, creating new collective works, for resale or redistribution to servers or lists, or reuse of any copyrighted component of this work in other works.
}
\endgroup

\begin{abstract}
We investigate how formal temporal-logic specifications can enhance the safety and robustness of reinforcement-learning (RL) control in aerospace applications. Using the open-source AeroBench F-16 simulation benchmark, we train a Proximal Policy Optimization (PPO) agent to regulate engine throttle and track commanded airspeed. The control objective is encoded as a Signal Temporal Logic (STL) requirement: maintain airspeed within a prescribed band during the final seconds of each maneuver. To enforce this specification at run time, we introduce a conformal STL shield that filters the RL agent’s actions using online conformal prediction. We compare three settings—(i) PPO baseline, (ii) PPO with a classical rule-based STL shield, and (iii) PPO with the proposed conformal shield—under both nominal conditions and a severe stress scenario involving aerodynamic-model mismatch, actuator-rate limits, measurement noise, and mid-episode setpoint jumps. Experiments show that the conformal shield preserves STL satisfaction while maintaining near-baseline performance and providing stronger robustness guarantees than the classical shield. These results demonstrate that combining formal specification monitoring with data-driven RL control can substantially improve the reliability of autonomous flight control in challenging environments.
\end{abstract}
\begin{keywords}
Reinforcement Learning, Signal Temporal Logic (STL), Conformal Prediction, Safe Control, F-16 Engine Simulation.
\end{keywords}
\section{Introduction}
\label{sec:intro}

Reinforcement learning (RL) has emerged as a powerful technique for controlling complex dynamical systems across a considerable range of domains—from robotics and autonomous driving to industrial process control.  
By learning control policies directly from data, RL can adapt to highly nonlinear dynamics that are often intractable for classical controllers\cite{groot2022improving}.  
In aerospace applications, for instance, RL can optimize maneuvering and adapt to rapidly changing flight conditions while satisfying performance objectives
\cite{clarke2020deep}.

The F--16 fighter aircraft provides an especially challenging and instructive case study.  
It was intentionally designed to be \emph{slightly aerodynamically unstable} in order to improve agility and maneuverability.
When the aircraft is not kept in a trimmed state by an active control law, 
small perturbations can grow into increasing oscillations.  
Recent incidents—such as the F-16 crash in Poland during the execution of a 
\textit{barrel roll maneuver}\cite{ASN541151}—underscore that even advanced aircraft with robust control systems remain vulnerable under unexpected failures or disturbances. 
While investigations of such crashes often involve many contributing factors (mechanical, environmental, human),
they highlight the need for control methods that can accommodate off-nominal conditions and provide quantitative safety guarantees. 

We adopt the open-source AeroBench F-16 benchmark~\cite{heidlauf2018verification},
a high-fidelity Python simulator with an inner-loop LQR and documented
outer-loop scenarios (engine control, longitudinal dynamics, ground-collision
avoidance). Its realistic aerodynamics and built-in safety specifications make it
well suited for evaluating our conformal STL–based safe-RL framework.
\paragraph*{Related Work.}
Safe-RL techniques—e.g., constrained optimization~\cite{andersson2015model}, reward shaping~\cite{qian2023reward}, or shielding~\cite{konighofer2023online}—reduce violations but usually lack formal probabilistic guarantees, especially under abrupt changes.
Formal verification tools such as \emph{Signal Temporal Logic} (STL) provide interpretable, quantitative measures of temporal-logic specification satisfaction.
Statistical model checking and predictive runtime verification have been used to assess STL properties of stochastic systems~\cite{lindemann2023conformal}.  
Recent work~\cite{zhao2023robust} applies conformal prediction (CP) to STL monitoring to obtain distribution–free, finite-sample confidence bounds on specification satisfaction, including under moderate distribution shift.  
However, these works focus on offline or purely model–based verification and do not address learning-based controllers.  
To the best of our knowledge, integrating conformal STL monitoring directly into the control loop of a reinforcement-learning agent has not yet been investigated.
Conformal prediction~\cite{shafer2008tutorial} provides distribution-free and informative bounds on prediction error.  
Recent robust CP methods preserve statistical coverage even when the test distribution deviates moderately from calibration.
Combining CP with STL monitoring enables a probabilistic safety filter~\cite{fan2020statistical}: short-horizon predictions are checked against STL specifications, and conformal quantile bounds guarantee that the probability of violation stays below a chosen risk level.
Several works have explored the intersection of RL and
STL.  
Venkataraman et al.~\cite{venkataraman2020tractable} encode STL tasks directly into the RL
objective to learn policies that satisfy given specifications, while
other approaches~\cite{kapoor2020model,singh2023stl}
focus on quantitative STL monitoring.
However, these methods have not considered  \emph{distribution–free}
finite–sample guarantees via conformal prediction on their proposed solutions.

\paragraph*{Contributions.}
This paper proposes a safe RL framework for F--16 flight control that integrates conformal STL monitoring with a runtime shield.
A Proximal Policy Optimization (PPO) policy, trained in the high-fidelity AeroBench F--16 simulator, provides control actions.
A short-horizon predictor, calibrated via robust CP, forecasts critical states such as airspeed and engine power.
Predicted trajectories are evaluated against STL safety formulas; when worst-case robustness becomes non-positive, the shield overrides the RL action with a safe recovery maneuver.
This preserves mission performance while ensuring probabilistic safety.

\section{Background}

\subsection{Reinforcement Learning}
RL\cite{wwiering2012reinforcement} 
models sequential decision making as a Markov
decision process (MDP) $(\mathcal{S},\mathcal{A},P,r,\gamma)$,
where $\mathcal{S}$ and $\mathcal{A}$ are the state and action spaces,
$P(s'|s,a)$ is the transition kernel, $r$ the reward, and
$\gamma\in[0,1)$ the discount factor.
At time $t$ the agent samples $a_t\!\sim\!\pi_\theta(\cdot|s_t)$,
receives reward $r_t$, and moves to $s_{t+1}$.
The objective is to maximize the discounted return
\begin{equation}
J(\theta)=\mathbb{E}\!\left[\sum_{t=0}^{\infty}\gamma^t r(s_t,a_t)\right].
\end{equation}
We employ \emph{Proximal Policy Optimization} (PPO) for its stability
and sample efficiency in continuous control~\cite{engstrom2019implementation}.

\subsection{Signal Temporal Logic (STL)}

STL is a formal language for specifying time–dependent
properties of real–valued signals \cite{deshmukh2017robust}.  
An STL formula $\varphi$ is generated by the grammar
\begin{equation*}
\varphi ::= \mu
\;\big|\;
\lnot \varphi
\;\big|\;
\varphi_1 \land \varphi_2
\;\big|\;
\mathbf{F}_{I}\varphi
\;\big|\;
\mathbf{G}_{I}\varphi ,
\end{equation*}
where $\mu$ is an \emph{atomic predicate} of the form $h(x_t)\ge 0$,
and $I\subseteq \mathbb{R}_{\ge 0}$ is a bounded time interval.
The temporal operators $\mathbf{F}_I$ (``eventually'') and
$\mathbf{G}_I$ (``always'') express that $\varphi$ must hold
at some or all times within $I$.

For example,
$\mathbf{G}_{[T-W,T]}\bigl(|(V_t-\mathrm{sp})/\mathrm{sp}|\le \tau\bigr)$
requires the airspeed error to remain within a tolerance $\tau$ during
the final $W$ seconds of the episode.
Beyond Boolean satisfaction,
STL provides a \emph{robust} semantics
$\rho^\varphi(x,t)\in\mathbb{R}$ that measures the signed distance
from violation: $\rho^\varphi(x,t)>0$ indicates that the property holds
with a margin, and $\rho^\varphi(x,t)<0$ indicates violation\cite{varnai2020robustness}.
This quantitative robustness is what our runtime monitor evaluates
when combined with conformal prediction in Section~\ref{sec:method}.

\subsection{Conformal Prediction}
Conformal prediction (CP) builds prediction sets $\mathcal{C}(x)$ for a target
$y$ such that
\begin{equation}
\mathbb{P}\bigl(y\in\mathcal{C}(x)\bigr)\ge 1-\delta
\end{equation}
for a user-specified miscoverage $\delta\in(0,1)$, without assumptions on the
data distribution.
In regression~\cite{johansson2018interpretable}, split CP calibrates a model $f$ on held-out data
$\mathcal{D}_{\text{cal}}$ to obtain residual quantiles
\begin{equation}
\hat q_{1-\delta}=
\operatorname{Quantile}_{1-\delta}\bigl(|y_i-f(x_i)|\bigr)_{(x_i,y_i)\in\mathcal{D}_{\text{cal}}}.
\end{equation}
Predictions satisfy $y\in [\,f(x)\pm \hat q_{1-\delta}\,]$ with $1-\delta$
coverage.

\subsection{Robust Conformal STL Monitoring}
Recent work~\cite{lindemann2023conformal} extends CP to moderate distribution shifts by inflating the calibrated quantiles using bounds on an $f$-divergence between the calibration and test distributions.
Under the assumption that the test distribution lies within an $f$-divergence ball around the calibration distribution, coverage can be preserved by enlarging the prediction sets accordingly
\cite{namkoong2016stochastic}
.
Given a predicted trajectory $\hat x_{t:t+H}$ with conformal bounds, the STL
robustness
\begin{equation}
\rho^\varphi_{\min}(t)=
\min_{x\in\mathcal{C}(\hat x_{t:t+H})}\rho^\varphi(x,t)
\end{equation}
represents the worst-case satisfaction.  A negative
$\rho^\varphi_{\min}(t)$ signals an imminent specification violation and
triggers a safety override.

\section{Methodology}\label{sec:method}
\subsection{Problem Setup}
We study safe outer–loop control for the F--16 \emph{engine} benchmark with continuous state
$s_t = [V_t,\,\text{pow}]^\top \in \mathbb{R}^2$ and scalar action $a_t \in [0,1]$ (throttle).
The objective is to track a commanded airspeed setpoint $\text{sp}$ while enforcing a run--time
STL constraint over the last portion of the episode. Let
$\rho^\varphi(x, t)$ as the robustness of trajectory $x$ with respect to STL formula $\varphi$ at time~$t$.
\FloatBarrier
\begin{figure}[!t]
  \centering
  \includegraphics[width=0.442
  \textwidth]{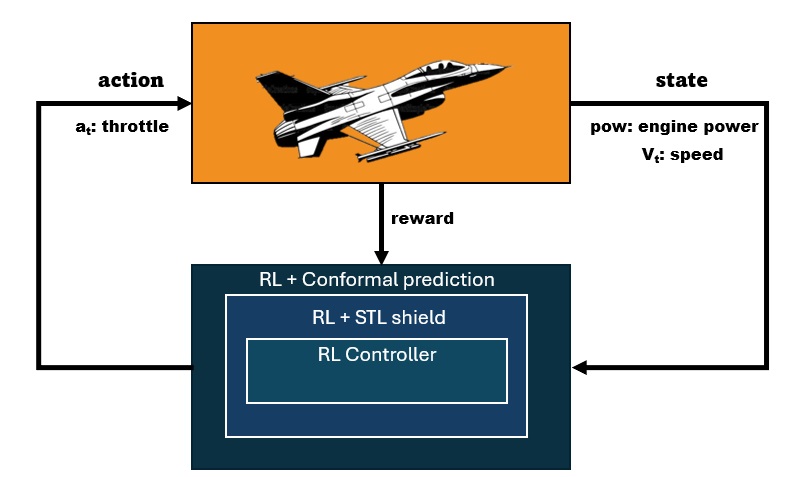}
  \vspace{-0.1em}
  \caption{Schema of the proposed solution}
  \label{fig:rl}
  \vspace{-0.1em}
\end{figure}

We wrap our proposed plant as a Gym/Gymnasium environment~\cite{towers2024gymnasium}.
As depicted in
Fig.~\ref{fig:rl},
the plant integrates only two states: true airspeed $V_t$ and an engine lag state \texttt{pow}; all other aircraft
states are held fixed to a straight--level trim. The observation is
$\big[V_t/\text{sp},\; \text{pow}/100\big]$; the action is throttle in $[0,1]$.
Unless stated otherwise, we use $dt{=}0.1$\,s, horizon $T{=}60$\,s, and a nominal setpoint $\text{sp}{=}500$\,ft/s.

\subsection{STL Safety Specification}
Our safety requirement is a bounded tracking error at the end of the episode. We use
\begin{equation}
\label{eq:stl}
\varphi_{\text{track}} \;:=\; \mathbf{G}_{[\,T-W,\,T\,]}\!\left(\;\big|\tfrac{V_t - \text{sp}}{\text{sp}}\big| \le \tau\;\right),
\end{equation}
with nominal $(\tau,W)=(0.05,\,10\text{ s})$ and a stricter stress case $(\tau,W)=(0.02,\,5\text{ s})$.
We evaluate the robustness $\rho^{\varphi_{\text{track}}}$ using a standard monitor over the observed trajectory.

\subsection{Policy and Training}
We train a PPO policy $\pi_\theta(a\mid s)$ that outputs throttle.
The reward combines tracking and smoothness,
\begin{equation}
r_t \;=\; -\Big|\tfrac{V_t-\text{sp}}{\text{sp}}\Big| \;-\; \alpha\,\Delta a_t^2 \;+\; \beta\,\!\left(\Big|\tfrac{V_t-\text{sp}}{\text{sp}}\Big|\le 0.05\right),
\end{equation}
We do not bake STL into the reward; STL is enforced at run time.
The coefficients 
$\alpha$ and 
$\beta$
were tuned empirically via a coarse grid search to achieve stable training and good trade-off between tracking performance and control smoothness.

\subsection{Runtime Shields}
We compare three execution modes:
\begin{enumerate} 
  \setlength{\itemsep}{1pt}
  \setlength{\parskip}{1pt}
  \setlength{\parsep}{1pt}
\item \textbf{PPO (baseline):} apply the raw policy throttle $u_t^{\text{RL}}$. \item \textbf{PPO+STL shield (rule--based):} pass $u_t^{\text{RL}}$ through a rate--limited filter $u_t=\text{clip}(u_{t-1}+\Delta u;\;|\Delta u|\le s)$ to reduce oscillations. \item \textbf{PPO+Conformal STL shield:} predict one–step $V_{t+1}$ with a linear model $\widehat{V}_{t+1}=a\,V_t + b\,\text{pow}_t + c\,u_t + d$, calibrated via split conformal prediction. Denote the $(1{-}\delta)$ residual quantile by $\hat{q}$. For a candidate throttle $u$ forming an uncertainty interval $[\widehat{V}_{t+1}\!\pm\!\hat{q}]$ and compute the \emph{worst--case} STL robustness over this interval (propagated over a short horizon $K$ by iterating the predictor). The shield chooses \begin{align} u_t &\in \arg\max_{u \in \mathcal{U}(u_{t-1})} \; \rho_{\min}^{\varphi_{\text{track}}}(t; u), \\ \mathcal{U}(u_{t-1}) &= \bigl\{\,u\in[0,1] : |u-u_{t-1}| \le s \bigr\}. \end{align} and applies the most permissive action that keeps $\rho_{\min}^{\varphi_{\text{track}}}\!>\!0$ when possible; otherwise it picks the action with the largest robustness margin. Here $s$ is the slew bound, maximum allowed change in throttle, and $K$ the look–ahead (we use $K{=}6$, $\delta{=}0.30$). \end{enumerate}

\subsection{Stress Scenario}
To probe robustness, we evaluate under a composite stress stack~\cite{mandrioli2023stress}: (i) aerodynamic model switch,
(ii) throttle cap $u\le u_{\max}$, (iii) action slew limit, (iv) observation noise and action delay, and
(v) introducing a mid–episode setpoint jump. For this case we also evaluate the stricter STL parameters $(\tau,W)=(0.02,\,5\text{ s})$.

\subsection{Algorithmic Summary}
\label{sec:algorithms}

To clarify how the proposed method operates in practice, we separate it into
two key procedures.  
The first procedure calibrates the one–step predictor and computes the
conformal quantile needed for statistical coverage.  
The second procedure describes the on-line controller that couples the PPO
policy with the conformal–STL shield during execution.
Together they define the complete proposed safe–RL pipeline.

\begin{algorithm}[t]
\caption{Calibration of the Conformal Predictor}
\label{alg:calib}
\begin{algorithmic}[1]
\STATE Collect rollouts of the PPO policy under nominal conditions
\STATE For each timestep, store $(V_t,\;\text{pow}_t,\;\text{throttle}_t)$
\STATE Fit the one–step predictor $\hat V_{t+1}=a V_t + b\,\text{pow}_t + c\,u_t + d$
\STATE Compute absolute residuals on a held–out calibration set
\STATE Set quantile bound $q_{1-\delta}$ (robustly inflated if desired)
\STATE Store $q_{1-\delta}$ for use by the runtime shield
\end{algorithmic}
\end{algorithm}

\begin{algorithm}[t]
\caption{Runtime Control with Conformal–STL Shield}
\label{alg:shield}
\begin{algorithmic}[1]
\STATE Initialize environment and set previous throttle $u_{-1}=0.5$
\FOR{each time step $t$ until episode end}
    \STATE RL proposes action $u_t^{\mathrm{RL}}$ from PPO policy
    \STATE Predict one–step $\widehat{V}_{t+1}$ and build interval
           $[\widehat{V}_{t+1}\!\pm\!q_{1-\delta}]$
    \STATE Compute worst–case STL robustness
           $\rho_{\min}^{\varphi_{\text{track}}}(t; u)$
    \IF{$\exists u$ within slew bound s.t. $\rho_{\min}>0$}
        \STATE Apply the most permissive such $u$
    \ELSE
        \STATE Apply the $u$ with largest robustness margin
    \ENDIF
\ENDFOR
\end{algorithmic}
\end{algorithm}

Algorithm~\ref{alg:calib}  collects trajectories of the trained PPO policy under nominal conditions,
fit a simple linear one-step model of airspeed, and determine the
$(1-\delta)$ conformal quantile of the prediction residuals.
This single scalar quantile provides a distribution-free bound on the
one-step prediction error.

Algorithm~\ref{alg:shield} gives the runtime procedure.
At each control step the PPO policy proposes a throttle command.
The calibrated predictor generates an interval forecast of the next airspeed
and the shield evaluates the worst-case STL robustness over that interval.
If there exists at least one throttle value within the allowed slew range
that preserves positive robustness, the shield applies the most permissive
such value; otherwise it selects the throttle with the highest robustness
margin.  
This ensures that the STL tracking specification is satisfied with the
targeted probabilistic coverage while leaving the RL controller as
unrestricted as safety permits.
\section{Results and Discussion}
\label{sec:results}
\begin{figure}[t]
  \centering
  \includegraphics[width=\columnwidth]{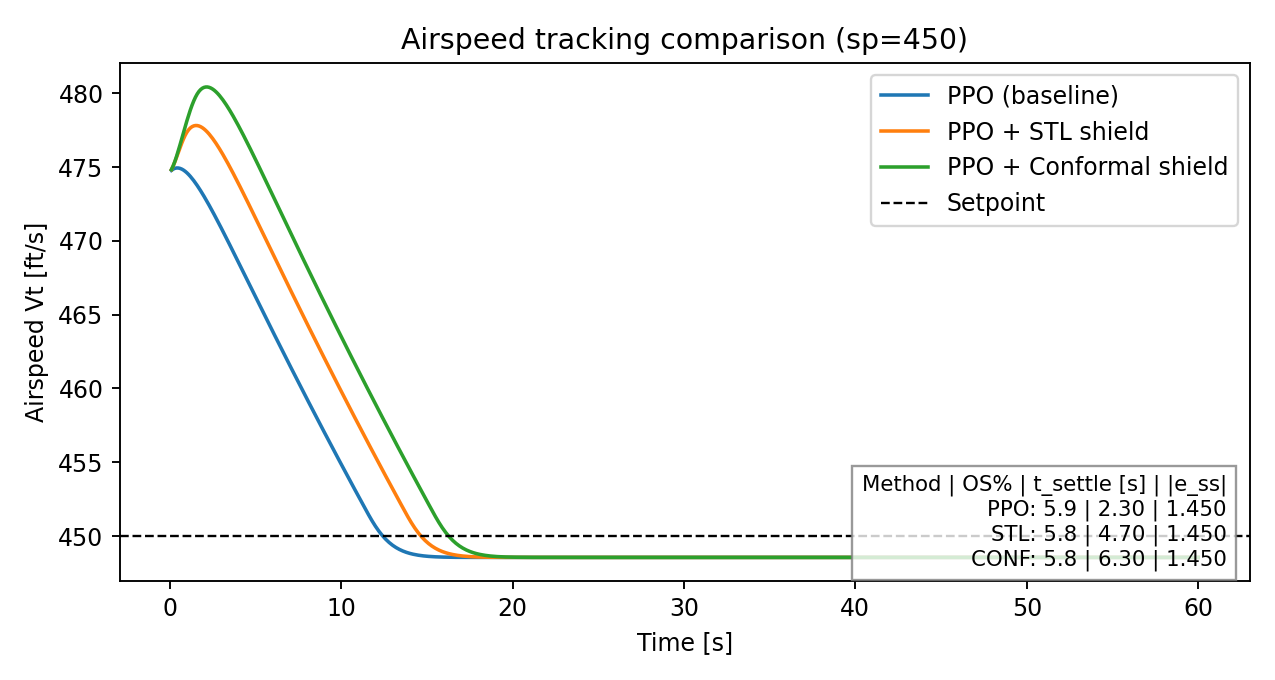}
  \vspace{-0.75em}
  \caption{Airspeed tracking with the shifted setpoint to $450$~ft/s.}
  \label{fig:nominal}
  \vspace{-0.1em}
\end{figure}

\noindent
\textbf{Nominal condition.}
Under nominal conditions ($\text{sp}{=}500$\,ft/s, $dt{=}0.1$\,s, $T{=}60$\,s), all three execution modes satisfy the STL spec with near-identical robustness. 
Figure~\ref{fig:nominal} shows the airspeed trajectories when the commanded
setpoint is shifted from $500$ to $450$~ft/s.
Under this nominal condition the three controllers behave almost similarly.
The plain PPO agent reaches the new setpoint fastest, with a settling time
of about $2.3$~s and overshoot of roughly $5.9\%$.
Adding the STL shield slightly delays convergence ($4.7$~s) while the
conformal STL shield introduces the largest delay ($6.3$~s), reflecting
its more conservative safety margin.
Despite these differences,
all three methods achieve virtually identical steady--state error
($|e_{ss}|\approx 1.45$~ft/s) maintaining the 
STL tracking specification.

\label{sec:stress}
\noindent
\textbf{Progressive stress conditions.}
To evaluate the generalization and safety of the trained controllers we designed a
\emph{progressive benchmark suite} of four scenarios.  
The policy is trained \emph{once} on the nominal plant and then tested
\emph{without retraining} as the environment is made increasingly adversarial.
Each introduces stricter STL requirements and/or
added disturbances:

\begin{itemize}
  \setlength{\itemsep}{3pt}
  \setlength{\parskip}{3pt}
  \setlength{\parsep}{3pt}
  
  \item \textbf{S0 – Nominal:} no disturbance; STL tracking specification
  $\varphi_{\text{track}} = \mathbf{G}_{[T-10,T]}\big(|(V_t-\mathrm{sp})/\mathrm{sp}| \le 0.05\big)$.
  \item \textbf{S1 – Mild:} small rate limit and low observation noise,
        STL tightened to $4\%$ band and $8$\,s final window.
  \item \textbf{S2 – Moderate:} throttle saturation and increased delay or noise,
        STL tightened to $3\%$ band and $6$\,s window.
  \item \textbf{S3 – Strong:} full stress stack— model mismatch (Morelli),
        throttle cap, stronger noise and delay, and a mid–episode setpoint jump;
        STL specification tightened to $2\%$ band and $5$\,s window.
\end{itemize}


For each scenario we execute $N=20$ independent rollouts per controller.
For each rollout the STL monitor computes the satisfaction outcome and the robustness value $\rho_i$, where $\rho_i>0$ denotes satisfaction and $\rho_i<0$ a violation.

While Table~\ref{tab:stress} reports only the mean robustness $\bar\rho$, STL satisfaction depends on the sign of individual robustness values.
Thus, controllers with similar $\bar\rho$ may exhibit very different satisfaction rates if their robustness distributions differ.
In particular, the Conformal STL shield reduces the occurrence of negative robustness events, which explains its higher satisfaction rate despite comparable mean robustness.

\begin{table}[t]
\centering
\caption{Progressive stress benchmark}
\label{tab:stress}
\begin{tabular}{l l r r}
\toprule
\textbf{Scenario} & \textbf{Method} & \textbf{Sat.\%} & \textbf{Mean $\bar\rho$} \\
\midrule
S0 Nominal (5\%/10s)  & PPO          & 100.0 & 0.044 \\
                      & PPO+STL      & 100.0 & 0.044 \\
                      & PPO+Conformal& 100.0 & 0.044 \\
\midrule
S1 Mild (4\%/8s)      & PPO          & 100.0 & 0.030 \\
                      & PPO+STL      & 100.0 & 0.032 \\
                      & PPO+Conformal& 100.0 & 0.032 \\
\midrule
S2 Moderate (3\%/6s)  & PPO          & 90.0 & 0.016 \\
                      & PPO+STL      & 95.0 & 0.021 \\
                      & PPO+Conformal& 100.0 & 0.020 \\
\midrule
S3 Strong (2\%/5s)    & PPO          & 60.0 & 0.006 \\
                      & PPO+STL      & 75.0 & 0.017 \\
                      & PPO+Conformal&  95.0 & 0.017 \\
\bottomrule
\end{tabular}
\end{table}

\noindent

\textbf{Discussion.}
Under nominal conditions (S0), all controllers satisfy the STL requirement with a large robustness margin.  
As the specification tightens and disturbances increase (S1–S3), the unshielded PPO controller degrades significantly, dropping to $60\%$ satisfaction in S3.  
Both shielding methods improve robustness, with the Conformal STL shield achieving $95\%$ satisfaction in the hardest scenario—over $30$ percentage points higher than PPO.  
This shows that conformal STL monitoring provides a strong probabilistic safety guarantee under distribution shift without retraining.

The additional runtime introduced by conformal monitoring was negligible in our experiments (below two seconds end-to-end).
This is mainly due to the low-dimensional state and action spaces of the benchmark.
For higher-dimensional systems, computational overhead may become significant, and we leave a detailed scalability analysis to future work.
The full implementation is available in our repository~\cite{repo}.

\bibliographystyle{IEEEbib}
\bibliography{strings,refs}

\end{document}